# VEHICLE DETECTION AND TRACKING TECHNIQUES: A CONCISE REVIEW

Raad Ahmed Hadi[1, 2], Ghazali Sulong[1] and Loay Edwar George[3]

[1]Department of Computer Science, University Technology Malaysia, Skudai, Malaysia
[2]Department of Software and Networking Engineering,
College of Engineering, Iraqi University, Baghdad, Iraq
[3]Department of Computer Science,
College of Science, Baghdad University, Baghdad, Iraq

## ABSTRACT

*Vehicle detection and tracking applications play an important role for civilian and military applications such as in highway traffic surveillance control, management and urban traffic planning. Vehicle detection process on road are used for vehicle tracking, counts, average speed of each individual vehicle, traffic analysis and vehicle categorizing objectives and may be implemented under different environments changes. In this review, we present a concise overview of image processing methods and analysis tools which used in building these previous mentioned applications that involved developing traffic surveillance systems. More precisely and in contrast with other reviews, we classified the processing methods under three categories for more clarification to explain the traffic systems.*

## 1. INTRODUCTION

One of the significant applications of video-based supervision systems is the traffic surveillance. So, for many years the researches have investigated in the Vision-Based Intelligent Transportation System (ITS), transportation planning and traffic engineering applications to extract useful and precise traffic information for traffic image analysis and traffic flow control like vehicle count, vehicle trajectory, vehicle tracking, vehicle flow, vehicle classification, traffic density, vehicle velocity, traffic lane changes, license plate recognition, etc. [1-4]. In the past, the vehicle detection, segmentation and tracking systems used to determine the charge for various kinds of vehicles for automation toll levy system [5]. Recently, vehicle recognition system is used to detect (the vehicles) or detect the traffic lanes [6-10] or classify the type of vehicle class on highway roads like cars, motorbikes, vans, heavy goods vehicles (HGVs), buses and etc. [5, 7, 11-15].

However, the traditional vehicle systems may be declines and not recognized well due to the vehicles are occluded by other vehicles or by background obstacles such as road signals, trees, weather conditions, and etc., and the performance of these systems depend on a good traffic image analysis approaches to detect, track and classify the vehicles.

                                    1

In this review paper, the traffic image analysis comprises of three parts: (1) Motion Vehicle Detection and Segmentation Approaches (2) Camera Calibration Approaches and (3) Vehicle Tracking Approaches.

The rest of the paper is organized as follows: motion vehicle detection and segmentation approaches are discussed in section 2. In section 3 the camera calibration approaches are illustrated. Vehicle tracking approached are presented in section 4. The other approaches summarized in section 5 and finally conclusions are summed up in section 6.

## 2. MOTION VEHICLE DETECTION AND SEGMENTATION APPROACHES

The detection of moving object's regions of change in the same image sequence which captured at different intervals is one of interested fields in computer vision. An important large number of applications in diverse disciplines are employed the change detection in its work, such as video surveillance, medical diagnosis and treatment, remote sensing, underwater sensing and civil infrastructure [16]. One of the video surveillance branches is the traffic image analysis which included the moving/motion vehicle detection and segmentation approaches. Even though various research papers have been showed for moving vehicle detection (background subtraction, frame differencing [17-22] and motion based methods) but still a tough task to detect and segment the vehicles in the dynamic scenes. It consists of three main approaches to detect and segment the vehicle, as mentioned below:

1. Background Subtraction Methods.

2. Feature Based Methods.

3. Frame Differencing and Motion Based methods.

### 2.1. Background Subtraction Methods

The process of extracting moving foreground objects (input image) from stored background image (static image) or generated background frame form image series (video) is called background subtraction, after that, the extracted information (moving objects) is resulted as the threshold of image differencing. This method is one of widely change detection methods used in vehicle regions detection. The non-adaptively is a drawback which is raised due to the changing in the lighting and the climate situations [23]. So, several researchers work to resolve this drawback by proposed methods on this field.

A significant contribution suggested the statistical and parametric based techniques which are used for background subtraction methods; some of these methods used the Gaussian probability distribution model for each pixel in the image [24-28]. After that, the pixel values updated by the Gaussian probability distribution model these pixel values which are updated from new image in the new image series. Then, each pixel (x,y) in the image is categorized either be a part of the foreground (moving object or called blobs) or background according to adequate amount of knowledge accumulated from the model which mention above using the equation (1) below:

$I(x, y) - Mean(x, y) < (C \times Std (x, y))$       (1)

Where $I(x, y)$ is pixel intensity, C is a constant, $Mean(x, y)$ is the mean, $Std (x, y)$ is the standard deviation.

An advanced background subtraction technique used to detect and extract features for vehicles in complex road scenes in traffic surveillance. This innovative technique uses a filtering method based on a histogram which collects information from sequences of frames of scatter background. This proposed background subtraction algorithm depicted a well performance under different conditions including various view-angles, overcrowding and illumination [29].

Another work proposed by [30] which is an example-based algorithm for detecting vehicles in traffic supervision video streams labeled the vehicles from examples for detection process. It involved from, firstly, an adaptive background approximation is used, then, dividing the image into small non overlapped blocks for founding the candidate vehicles parts from these blocks. Secondly, Principal Component Analysis (PCA) is applied as a low-dimensional statistical method to measure the two histograms of each candidate, and support vector machine (SVM) is considered for real vehicle parts classification. Eventually, all classified parts shaped and connected as a parallelogram to represent the parts shapes for matching process.

Also, a new method for vehicle detection based on shadows underneath vehicles information has proposed by [9]. This method extracts the size features of vehicles from information that gathered form the distance between ends of front and rear tires for underneath shadow of vehicles to distinguish the existence of vehicles on the lanes. In this paper, the information represented as traffic movement images which obtained from a camera assembled on a low position such as the roadside, sidewalk, and etc. Moreover, this method has accurate vehicle detection because it is used the functions to generate and improve a background image, in addition, approximate and modernize the value of threshold of background subtraction images binarization automatically (Figure 1).

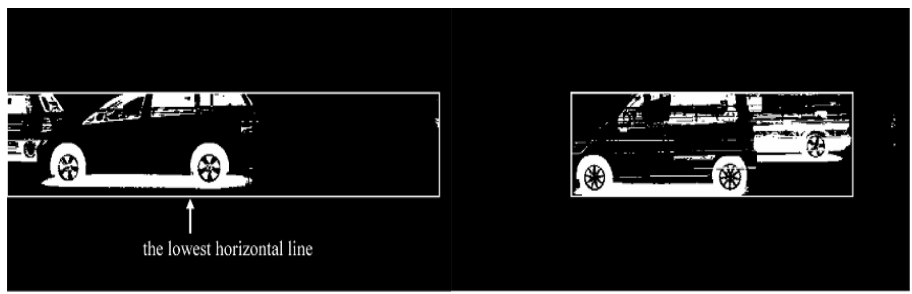

Figure (1) Shows the underneath shadow (a) the measurement area and the lowest horizontal line (b) A region of vehicles and their shadows.

Also, a novel well-organized idea based on suggested filters which are used for mobile vehicles detection have presented by [31]. In this research, the two filters are used to eliminate swinging trees and raindrops from forefront entities respectively, and the swinging trees filter is used to decrease the calculation difficulty of consequent vehicle tracing. In addition, a shade removal method is combined with a versatile background deletion approach to take out the mobile vehicles in background images.

The authors [32] presented a new idea for tracking vehicles by integrating these traits such as volume, location, color dissemination, speed of a group of the forefront entity and Gaussian Mixture based background form. In this approach, each pixel in the image view is demonstrated and categorized as either a noise or forefront entity's background. In addition, the author is employed a projective floor-level transform to reinforce the expectations of speed persistence and entity volume for forefront form.

## 2.2. Feature Based Methods

Another trend which the researchers investigate and motivate on sub-features like the edges and corners of vehicles, the moving objects segmented from background image by collecting and analyzing the set of these features from the movement between the subsequent frames. Furthermore, the feature based method supports the occlusion handling between the overlapping vehicles and compared with background subtraction method represents a less level from the computational difficulty view [33].

Several approaches can discriminate the object from the background by using its features, a trainable object detection approach has proposed by [34]. This approach based on learning which employs a set of labeled training data which used for labeling the extracted objects features. In addition, it uses a Haar wavelets technique as feature extraction method and also uses support vector machine classifier for classification process. Moreover, face, people and cars static images datasets have tested on this approach.

A subregion is a technique used to locate the local features which used for recognition non-occluded and partially occluded vehicles. Principal components analysis (PCA) weight vector used to pattern the low-frequency components and an independent component analysis (ICA) coefficient vector used to pattern the high-frequency components, these two vectors were generating by subregions. This approach represents a novel statistical method which dependent on local features of three subregions for detecting the vehicles automatically [35].

Furthermore, a multiscale transformation uses the frame elements of image which are indexed by position, measure and orientation criterions, and have time-frequency localization properties of wavelets also it shows a very high degree of directionality and anisotropy, this method called The curvelet transform. The curvelet transform used within a new vehicle recognition algorithm as a feature extraction method has offered by [15]. The authors presents that there are three various types of classifiers are used in this paper for vehicle recognition: k nearest-neighbor, Support vector machine (one versus one) and Support vector machine (one versus all). Finally, the vehicle recognition process showed a high performance through the experiments results.

A local-feature point's configuration method used for vehicle classification with using computer graphics (CG) model images has introduced by [36]. In this work, the eigen-window approach is used due to it has several advantages such as  detect the vehicles even if it changed its path due to veering out of the lanes and also if parts of the vehicles are occluded. In addition, the CG model images achieved a high performance results for vehicles recognition process to real images of vehicles. Furthermore, the CG model facilitates the task of collecting real images of all target vehicles because it is a time consuming and difficult task (Figure 2).

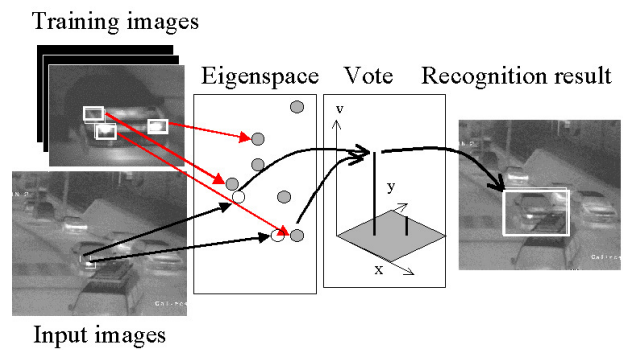

Figure (2) Eigen-window method.

Also, a new traffic criterion detection approach based on Epi-polar Plane Image (EPI) has proposed by [37]. This method treats the noise sensitivity and existence of the rough edge on edge detection through developed a new sobel operator which overcomes the traditional sobel shortcomings, and the Gabor operator texture edge detection is also used for extracting the features. Experimental tests examined that this approach is accurate and anti-nose edge detection.

In this paper [38] the authors have suggested a low resolution aerial image used as dataset for detection vehicles system, this system uses the edges of the car body, the edges of the front windshield and the shade as the features for the similarity process. The gathered extracted features knowledge is shaped in the structure of the Bayesian network that will use for integration of all features. In this research, experiments present good results even if tested images were more complicated.

## 2.3. Frame Differencing and Motion Based Methods

The frame differencing is the process of subtracting two subsequent frames in image series to segment the foreground object (moving object) from the background frame image. Also, the motion segmentation process is another fundamental step in detecting vehicle in image series which is done by isolating the moving objects (blobs) through analyzed and assignment sets of pixels to different classes of objects which based on orientations and speed of their movements from the background of the motion scene image sequence [16, 23, 39, 40].

An intraframe, interframe and tracking levels are suggested framework to recognize and manipulate occlusion vehicles. This paper showed by quantitative evaluation that the interframe and interframe could be used to manage and manipulate mostly of partial occlusions images, and tracking level could be used to manage and manipulate full occlusions images effectively [2].

A multimodal temporal panorama (MTP) method for real time vehicle detection and reconstruction have suggested by [41]. This method accurately used a remote multimodal (audio/video) monitoring system to extract and reconstruct vehicles in real-time motion scenes. A multimodal approach in addition to detection and motion estimations has helped during the reconstruction process of vehicles, which removed the occlusion, motion blurring and differences in perspective views.

Visual-based dimensional approximation is an approach that used to extract motion vehicles from traffic image series and adjust them with a simple disfigured vehicle pattern. In this approach, shadow removing technique is used, in addition to, the experimental tests show an effective performance and sufficient accuracy for general vehicle type classification within the approach mentioned above works on traffic vehicles motion images [5].

A new method based on versatile movement histogram technique for detection of moving vehicles have introduced by [40]. In this method, two procedures involved to segment and detect the vehicles in video sequence. The first step, a novel background changing method will use for bright changing in video scene. The second step, adaptable movement histogram-based vehicles detection is used, supported and modernized corresponding with movement histogram in the dynamic view.

Nighttime traffic supervision is a new trend in the traffic surveillance systems. A suggested real-time several vehicle detection and tracking approach based on motion vehicles in nighttime traffic view used the image segmentation and pattern analysis method to detect and identify the vehicles form its headlights and taillights. A multilevel histogram thresholding technique is applied to extract and bring bright objects of interest from nighttime road scene. Finally, this approach showed an effectively, robustly and feasibly results when experimented in different nighttime situations for traffic supervision vehicles recognition and identification [42].

## 3. CAMERA CALIBRATION APPROACEHS

Measuring the vehicle speed and precision of vehicle tracking methods rely on well camera calibration performance, and the camera calibration setup may be done in semi-automatically matter or by hand. Camera calibration is a vital procedure for well video-based surveillance systems [4].

A new automatic method for segmenting and tracking vehicles applied on a video taken by camera at low angle level relatively to the ground on highway road [3]. In this paper, expectation of high features is calculated by joining of region-based grouping procedure, background subtraction, projective transformation and using of plumb line projection (PLP).

A novel automatic traffic system which uses 2D spatio-temporal images has suggested by [6]. This system uses a TV camera to keep track of vehicles for the highway traffic within two slice windows for each traffic lane. The purpose of this system was classifying the passing vehicles by using these 3D measurements (height, width and length) in addition to count the vehicles and assessment their speeds. Also, this system showed a robust performance when it tested under different light situations involving vehicles lights at night and shades in the day (Figure 3).

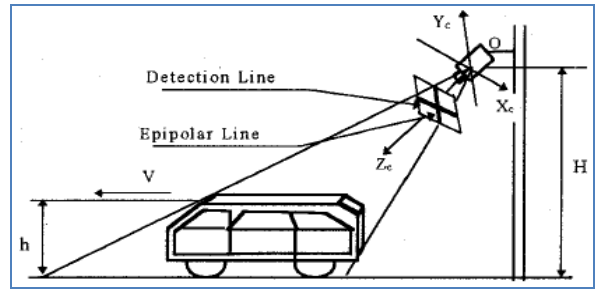

Figure (3) Camera setting

## 4. VEHICLE TRACKING APPROACHES

The object tracking in video processing is an important step to tracking the moving objects in visual-based surveillance systems and represents a challenging task for researchers [43]. To track the physical appearance of moving objects such as the vehicles and identify it in dynamic scene, it has to locate the position, estimate the motion of these blobs and follow these movements between two of consecutive frames in video scene [44]. Several vehicle tracking methods have been illustrated and proposed by several researchers for different issues, it consists of:

1. Region-Based Tracking Methods
2. Contour Tracking Methods
3. 3D Model-Based Tracking Methods
4. Feature-Based Tracking Methods
5. Color and Pattern-Based Methods

### 4.1. Region-Based Tracking Methods

In these methods, the regions of the moving objects (blobs) are tracked and used for tracking the vehicles. These regions are segmented from the subtracting process between the input frame image and prior stored background image.

A proposed research paper introduced a model-based automobile recognizing, tracking and classification which is efficiently working under most conditions [17]. The model provided position and speed knowledge for each vehicle as long as it is visible, in addition, this model worked on series of traffic scenes recorded by a stable camera for automobiles monocular images. The processing algorithms of this model represented of three levels: raw images, region level, and vehicle level.

A traffic criterions assessment such as vehicles numbering and classification involving with a suggested traffic observation scheme have suggested by [45]. The proposed scheme demonstrated in its work the feature ratio and density to classify vehicles, also, it used the geometric traits to eliminate the false regions and for more accurate segmentation process is used the shades elimination algorithm (Figure 4). Finally, this scheme experimented under three different lighting videos streams.

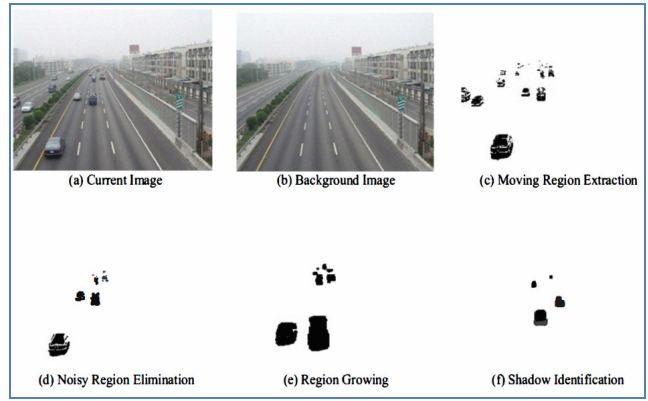

Figure (4) Detection and tracking of moving regions

## 4.2. Contour Tracking Methods

These methods depend on contours (the boundaries of vehicle) of vehicle in tracking vehicle process [39]. The authors [46]have proposed a novel real time traffic supervision approach which employs optical movement and uncalibrated camera parameter knowledge to detect a vehicle pose in the 3D world. In this paper, the proposed approach uses two new techniques: color contour based matching and gradient based matching, and it showed well results when it tested for tracking, foreground object detection, vehicle recognition and vehicle speed assessment methods.

A real-time vehicles tracking and classification technique on highway have offered by [8]. A few traffic criterions (lane change recognition, vehicle numbering and vehicle classification) are extracted by above technique. In addition, the proposed technique supported the occlusion detection and tracking which caused from multiple vehicles poses in the crowding situation. The proposed work used the Kalman filter, background differencing methods and morphological operations for extraction and recognition vehicle's contour.

## 4.3. 3D Model-Based Tracking Methods

The authors [47] have presented an occlusion detection approach based on generalized deformable model. In this paper, the occlusion of vehicles detection process used a 3D solid cuboid form with up to six vertices, and this cuboid used to fit any different types and sizes of vehicle images by changing the vertices for a best fit. Therefore, vehicle detection, segmentation and tracking can be achieved efficiently due to changes in the region proportion, prototype width and height with consideration to previous images.

A unified multi-vehicle tracking and categorization system for various types of vehicles such as motorcycles, cars, light trucks and heavy trucks on highway and windy road video sequences has recommended by [48]. In this paper, a vehicle anisotropic distance measurement achieved through the 3D geometric shape of vehicles.

A new 3D model-based vehicle detection and depiction framework is based on a probabilistic boundary feature grouping, which used for vehicle detection and tracking process has suggested by [49]. This framework supported advantageous traits such as the flexibility due to it is much free from the scale problem and detects the vehicles from more diagonal visions, in addition, it is fast when applied to many other applications.

## 4.4. Feature-Based Tracking Methods

An iterative and distinguishable framework based on edge points and modified SIFT descriptors as features uses in similarity process, these features represents a large region of set of features forms a strong depiction for object classes. The proposed framework showed a good performance for vehicle classification in surveillance videos despite of significant challenges such as limited image size and quality and large intra-class dissimilarities [50].

A linearity feature technique is a proposed line-based shade method which uses lines groups to remove all undesirable shades. The proposed method undertakes well the occlusion resulting from shades. Finally, this method represented an automatic vehicle tracking and classification traffic observation system [51].

An automatic unique visual-based expressway surveillance approach for segmenting and tracking vehicles during the image series with existence rigorous occlusion due to low-level floor position of camera on the roadside [52]. In this paper, the particular vehicles are detected, segmented and tracked in image sequence by assembling, bunching and approximating of the 3D world coordinates of vehicle's feature points.

## 4.5. Color and Pattern-Based Tracking Methods

An analyzing color of image series of traffic supervision views is a technique used by [53]. The authors uses the above technique by embraced the YCrCb color space for the construction preliminary background, segmenting foreground, vehicle location, vehicle tracking, shade elimination, and background updating algorithms that used through the proposed system. Through the practical experiments, this system proven could be to work under several weather situations, and it is insensitive to lighting.

A model-based system for real-time traffic supervision continuous visual tracing and classification of vehicles for busy multi-lane highway scene have recommended by [54]. In this proposed work, the authors use the orthographic approximations for matching process. This system consists of three improved main levels: (1) using 1-D patterns of shape and posture theory (2) theory tracking (3) using 2-D patterns of theory verification.

## 5. OTHER APPROACHES

Car-Rec is a real time car recognition system, this system used to detect cars in moving and steady matter. It used and combined multiple detection and feature descriptor extraction algorithms such as Speed-Up Robust Features (SURF) algorithm. This system tested on a large database of toy's car pictures. In addition, on of advantage Car-Rec framework is the modularity; which means this framework may be capable of development by modified the stages of this framework by replace one of image recognition algorithm by other and to modify this framework to do some of tasks [55].

In addition, the author [56] has suggested an approach to trace the trucks and automobile from overhead images view which called a shape-first approach. This approach tested on Overhead Imaging Research Dataset (OIRDS). The tracing process is utilizing the information from overhead image of an automobile, and recognizing the candidate vehicles from outlines of their shapes.

Also, the authors [57] have presented a vehicle segmentation method to detect, track and classify moving vehicles in presence of occlusion in crowded scenario. The authors used a vehicle outline patterns, camera adjustment and ground plane information for detection process instead of blob tracking methods due to unable to separate vehicles under occlusion, In addition, Markov Chain Monte Carlo (DDMCMC) process used within detection process.

On-road vehicle detection novel computational system uses Bayesian Network has proposed by [58]. Bayesian Network Enhanced Cascades Classification (BNECC) system utilized from texture and geometric features of vehicles during the recognition process. BNECC included on two parts of classifiers: the first one is a Cascade of Boosted Ensembles (CoBE) which informs about the vehicle texture features, and a Bayesian network that informs about volumes, position and confidence values of vehicle generated by CoBE.

The neural network employed and represented an important role for many applications such as image processing. Another significant contribution offered by [59], they have offered a neural-boundary-based vehicle detection and classification approach. The characteristics knowledge of traffic criterions such as, vehicle velocity, vehicle numbering, and vehicle category are taking out by merging the outcome of movement edge detection and background removal, this characteristic knowledge mining method is called seed-filling. Thereafter, the mined characteristic used as an entered data of neural network for an accurate vehicle recognition and categorization vehicle process.

Also a developed vehicle recognition method for traffic view explanation which is based on fuzzy integrals has offered by [60]. This algorithm uses and calculates the fuzzy integral which based on evidence gathered for recognition process, also, it extracts the contour boundaries of the vehicles by using the Hough transform and decreases the noise and enriches the form of the entity regions by using the morphological operations.

A detection of vehicle from front-vision is discussed by [61]. The authors have proposed a new method based on Markov chain Monte Carlo (MCMC) and using of maximizing a posterior probability (MAP) techniques. The Markov chain is planned to instances suggestions of edge knowledge from models of roads and vehicles, after that, consecutively vehicles detection process is done by using of MAP technique in front-view static images with frequent occlusions. Furthermore, a background subtraction and shade removing are not included and required as complex preprocessing steps for segmentation procedure. A lot of statistical approached used also in image processing.

Here, [62] have presented a computerized highway surveillance scheme based on higher order statistics (HOS) for classification of detecting vehicles in images. In this research, the classification is done by the learning of HOS information of the vehicle class from sample images, which is used as decision measure to classify test patterns as vehicles.

Moreover, [63] have suggested mobile vehicle segmentation method based on Gaussian gesture model. The Gaussian gesture model is used to shape, exploring and distinguishing between the mobile vectors of vehicles which are clustered in a small area and the mobile vectors of dynamic background which are scattered. Thereafter, the mobile in the views would be classifying by using the Bayesian system which supports the strength and enhancement for the proposed method and approximates the criterions of this method by using expectation–maximization (EM) algorithm.

## 6. CONCLUSIONS

This paper provides a summarizing study on the proposed techniques which have used in traffic video. It focuses in these areas, namely vehicle detection, tracking, and classification with appearance of shadow and partial occlusion. Also, we present and classify the traffic surveillance systems to three types based on specific methods which used for developing it. These types shows the detailed information about how the traffic surveillance systems used the image processing methods and analysis tools for detect, segment, and track the vehicles. In addition, shadow and partial occlusion matters and its available solutions are discussed. More specifically, this review gives better understanding and highlights the issues and its solutions for traffic surveillance systems.

## AUTHORS

**Raad Ahmed Hadi** was born in June 1, 1979, Baghdad, Iraq. He received his M.Sc. Computer Science 2006 Iraqi Commission for Computers and Informatics, Baghdad, Iraq, B.Sc. Computer Science 2003 University of Technology, Baghdad, Iraq. Currently he is PhD. student in University Technology Malaysia (UTM), Johor, Malaysia.

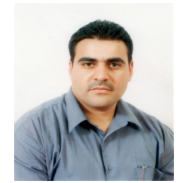

**Ghazali Sulong** was born in May 21, 1958 Malaysia. He received his Ph.D. Computing 1989 University of Wales College of Cardiff (UWCC), Wales, U.K., M.Sc. Computing 1982 University of Wales College Cardiff (UCC), Wales, U.K., B.Sc. Statistic 1979 UKM, Malaysia. Currently he is a Professor in Image Processing.

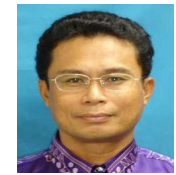

**Loay Edwar George** was born in July 1, 1957 Baghdad, Iraq. He received his Ph.D. Computer Science 1997 Baghdad University, Baghdad, Iraq, M.Sc. Physics 1983 Baghdad University, Baghdad, Iraq, B.Sc. Physics 1979 Baghdad University, Baghdad, Iraq. Currently he is a Professor in Image Processing.

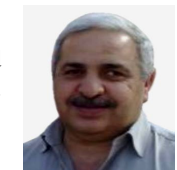